\documentclass[letterpaper ,conference]{IEEEtran}
\IEEEoverridecommandlockouts
\usepackage{cite}
\usepackage{amsmath,amssymb,amsfonts}
\usepackage{algorithmic}
\usepackage{graphicx}
\usepackage{textcomp}
\usepackage{color}
\usepackage{tabularx}
\usepackage{booktabs}
\usepackage[belowskip=0pt,aboveskip=0pt]{caption}
\newcommand{\fig}[1]{Fig.\ref{#1}}

\def\BibTeX{{\rm B\kern-.05em{\sc i\kern-.025em b}\kern-.08em
    T\kern-.1667em\lower.7ex\hbox{E}\kern-.125emX}}
\begin{document}

\title{Leveraging Pre-Trained 3D Object Detection Models For Fast Ground Truth Generation}

\author{\IEEEauthorblockN{Jungwook Lee, Sean Walsh, Ali Harakeh, and Steven L. Waslander}
\IEEEauthorblockA{\textit{Mechanical and Mechatronics Engineering} \\
\textit{University of Waterloo}\\
Waterloo, ON, Canada \\
j343lee@uwaterloo.ca, swalsh@uwaterloo.ca, www.aharakeh.com, stevenw@uwaterloo.ca}
}

\maketitle

\begin{abstract}
Training 3D object detectors for autonomous driving has been limited to small datasets due to the effort required to generate annotations. Reducing both task complexity and the amount of task switching done by annotators is key to reducing the effort and time required to generate 3D bounding box annotations.  This paper introduces a novel ground truth generation method that combines human supervision with pre-trained neural networks to generate per-instance 3D point cloud segmentation, 3D bounding boxes, and class annotations. The annotators provide object anchor clicks which behave as a seed to generate instance segmentation results in 3D. The points belonging to each instance are then used to regress object centroids, bounding box dimensions, and object orientation. Our proposed annotation scheme requires $30 \times$ lower human annotation time. We use the KITTI 3D object detection dataset \cite{geiger2012we} to evaluate the efficiency and the quality of our annotation scheme. We also test the the proposed scheme on previously unseen data from the Autonomoose self-driving vehicle to demonstrate generalization capabilities of the network.
\end{abstract}


\section{Introduction}
When looking at standard object detection benchmarks, there is a substantial discrepancy between the performance of 3D object detectors when compared to detectors in 2D. This can be attributed to the limited size and the low number of 3D object detection datasets. When available, these datasets \cite{geiger2012we, song2015sun} remain orders of magnitude smaller than their 2D object detection counterparts \cite{deng2009imagenet, everingham2010pascal}. Such small datasets are not sufficient to train high-capacity models that need to capture the additional complexity induced by adding a third dimension to the estimation problem.

\begin{figure}[t] 
\begin{center}
\includegraphics[width=\columnwidth]{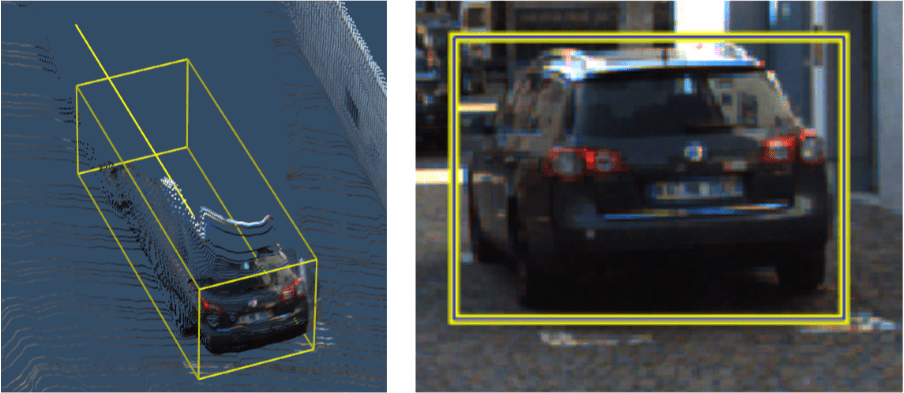}
\end{center}
\caption{Ground truth annotated bounding boxes from the KITTI \cite{geiger2012we} object detection benchmark. The vehicle can only be observed from a single view point in the 3D LIDAR point cloud, rendering the estimation of its extent difficult and time consuming for annotators.} 
\label{gt_kitti}
\end{figure}

The scarcity of labeled 3D detection data can be attributed to the time that is required for annotating a bounding box in 3D. Unlike 2D bounding box annotation, 3D bounding boxes have an additional dimension for annotators to handle and must be \textit{oriented} with respect to the sensor's coordinate frame. Furthermore, the data available to annotators from range scanners is sparse and 2.5 D \cite{vosselman2004recognising}, capturing surface information only from the sensor's viewpoint. \fig{gt_kitti} shows an example of an object from the \textit{KITTI object detection benchmark} \cite{geiger2012we} with ground truth annotations in 2D and in 3D. It can be seen that the object, in this case a vehicle, is only partially covered by the range scan. Estimating the 3D extent of the object from this 2.5 D data is a non-trivial inference task and requires additional annotator time to complete.

Why does it take so long to annotate a 3D bounding box? A typical 3D bounding box annotation procedure for indoor scenes is described in \cite{song2015sun}. First, annotators draw an imaginary 2D bounding box on the projection of the point cloud to the bird's eye view. This is followed by drawing an orientation arrow, still in the bird's eye view. Finally, the 2D bounding box is extruded to 3D by adjusting the top and bottom. This procedure is suboptimal for the task at hand as the annotators are required to switch between 3 separate tasks. Each task requires the annotators to divert their attention towards different parts of the objects, and requires them to use the mouse and keyboard in different ways. In cognitive psychology, the above procedure can be described as involving significant \textit{task switching}. It takes more time to complete tasks if a human must switch between them than if each task is completed for all instances before switching to the next \cite{gilbert2002task}. Furthermore, humans tend to exhibit higher error rates when task switching, when compared to performing one task at a time \cite{monsell2003task}. Also, task switching is cognitively demanding, resulting in mental fatigue after multiple iterations, which in turn decreases the annotators' performance \cite{van2003mental}. Finally, a substantial amount of mental imagery is required to estimate the extent of objects from 2.5D data. Similar to task switching, mental imagery has a high cognitive cost that substantially slows down the annotation task \cite{papadopoulos2017extreme}.

\begin{figure*}[t] 
\begin{center}
\includegraphics[width=\textwidth]{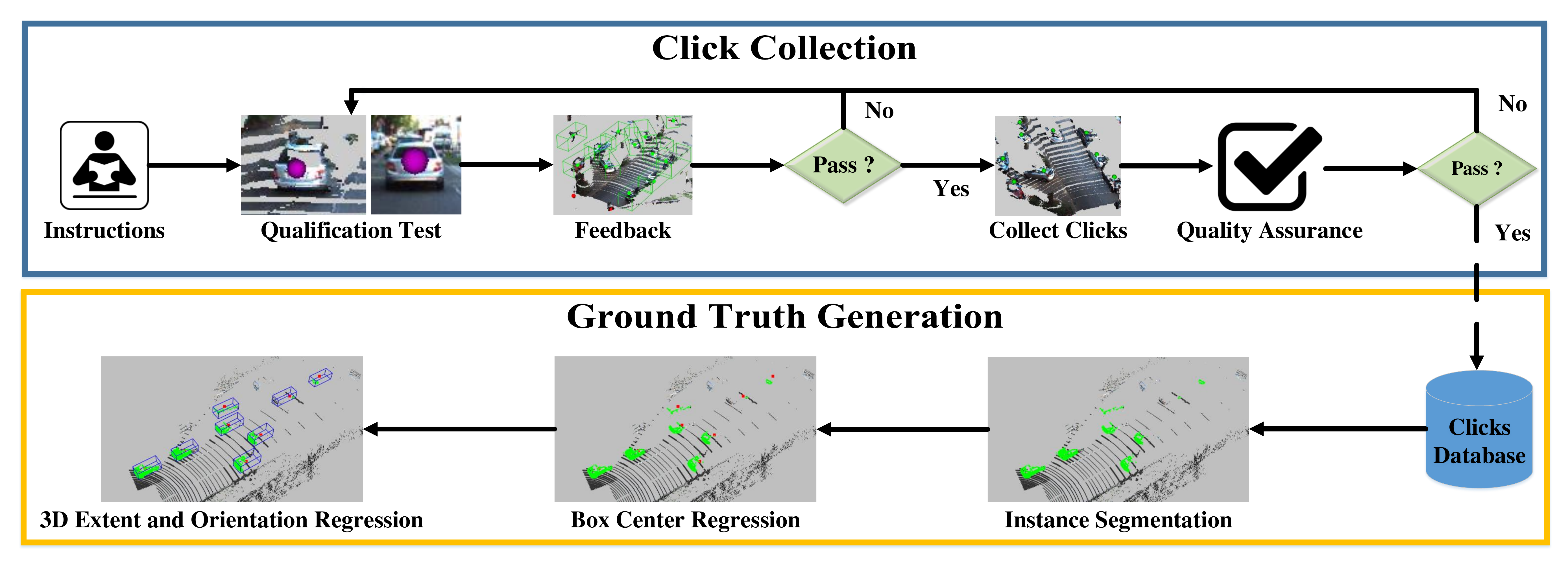}
\end{center}
\caption{Label3D architectural diagram. \textbf{Top:} The first stage is used to collect clicks representing points on the object. \textbf{Bottom:} The second stage uses the collected clicks to generate 3D ground truth bounding boxes by employing a pre-trained object detector to perform the 3D bounding box estimation task. The two tasks are independent and can be performed by two separate subsets of annotators.} 
\label{figure:algo}
\end{figure*}

This paper proposes a hybrid annotation scheme (\fig{figure:gui}) that consolidates the strengths of human annotators and deep 3D object detectors to quickly and efficiently generate high quality 3D detection ground truth labels. Similar to \cite{papadopoulos2017training} and \cite{papadopoulos2017extreme} for 2D, we ask annotators to \textit{click} on objects of a specific class in a LIDAR point cloud, using just one click per object. The provided clicks define segmentation seeds used to generate point cloud amodal instance level segmentation through a deep neural network. The segmentation results are finally provided to an amodal 3D bounding box estimation network to produce the bounding box results. The resultant framework eliminates task switching by assigning annotators the single task of clicking on each object in the scene. It also discards the need to determine the extent of a bounding box in 3D, reducing the task's mental imagery requirements. By performing extensive experiments on the KITTI dataset and our own novel dataset, \cite{geiger2012we} we demonstrate that:
\begin{itemize}
\item The proposed click annotation procedure is $30\times$ faster than the 3D bounding box annotation procedure described in \cite{song2015sun}, taking on average only $3.7$ seconds per object instance.
\item Leveraging pre-trained deep models to perform 3D instance level segmentation and 3D bounding box estimation leads to the generation of high quality ground truth labels, without any human annotations.
\item As a by-product of the process, each 3D bounding box is also associated with an instance level 3D segmentation mask.
\item Our proposed method can generalize to new data, thus is able to be used to label scenes outside the data it has been trained on.
\end{itemize}

\section{Related Work} 
\subsection{Time Required To Draw A Bounding Box}
Su et al. \cite{su2012crowdsourcing} is used in \cite{papadopoulos2016we, papadopoulos2017extreme, papadopoulos2017training} as a reference for the time required to draw a single 2D bounding box, at around 35 seconds. To our knowledge, no such reference exists for 3D bounding boxes. The KITTI dataset \cite{geiger2012we} does not report the time required to annotate its object detection benchmark\footnote{The authors of the KITTI dataset were contacted and they reported that such data is not kept in their records.}. The SUN RGB-D dataset\cite{song2015sun} reported that $2,051$ hours of annotation were required to label $64,595$ 3D object instances, around $114$ seconds per instance. We use this reported time for comparison in Section \ref{exp}.

\subsection{Click Supervision}
Click supervision has been previously used as a mechanism to reduce bounding box annotation time. \cite{papadopoulos2017training} used click supervision by incorporating clicks collected on the center of objects in 2D into a weakly supervised multiple instance learning scheme. This scheme provides an $18\times$ reduction in the time required for the 2D labelling task, but results in a slightly worse performing detector than one trained on human labelled data. \cite{papadopoulos2017extreme} uses click supervision to collect clicks at the four extremes of an object in a 2D image to deduce 2D bounding boxes. The resultant bounding boxes have a $97\%$ mean average precision when compared with human drawn ones on the PASCAL VOC dataset \cite{everingham2010pascal}, but require $5\times$ less time to annotate. 

Extending these two methods to 3D is a non-trivial task. Seeing the object from a single viewpoint in a range scan(\fig{gt_kitti}) makes it difficult for human beings to estimate its center or its extreme points. Our proposed scheme remedies this limitation by collecting clicks belonging to \textit{any surface point} on the object to be used as a seed for 3D instance segmentation, easing the task for annotators. Our proposed clicking scheme takes $3.7$ seconds per click \textit{in 3D} as opposed to $1.5$ seconds per click for both \cite{papadopoulos2017training,papadopoulos2017extreme} \textit{in 2D}. This increase in time is attributed to the additional time needed for visual search\cite{ehinger2009modelling} in a 3D point cloud, as well as the requirement that our annotators have to click on \textit{all} the instances in a frame as opposed to a \textit{single} instance in \cite{papadopoulos2017training,papadopoulos2017extreme}.

\subsection{Alternative Methods to Reduce Annotation Effort}
Alternative methods have been explored to reduce the annotation effort for the 2D object detection task. Weakly supervised object localization (WSOL) \cite{bilen2016weakly} has been used to train detectors using image level labels, a very cheap annotation technique. The resulting 2D object detectors are very weak compared to ones trained on human labelled bounding boxes achieving half the latter's mean average precision. Other works focus on getting the most out of as little training data as possible. Few shot learning was explored in \cite{dong2017few} to train 2D object detectors from as few as four samples per category. Similar to WSOL, the resulting detectors have yet to achieve comparable performance with detectors trained from human labels.

\section{The Proposed Annotation Scheme}
An overview of our annotation scheme can be seen in Figure \ref{figure:algo}. The first stage consists of collecting clicks from annotators using a custom designed 3D annotation tool. The second stage uses the collected clicks to generate instance level object segmentation followed by 3D bounding box estimation. The final output is a 3D bounding box and instance segmentation mask per collected click.

\subsection{Click Collection}
\begin{figure*}[t] 
\begin{center}
\includegraphics[width=0.95\textwidth]{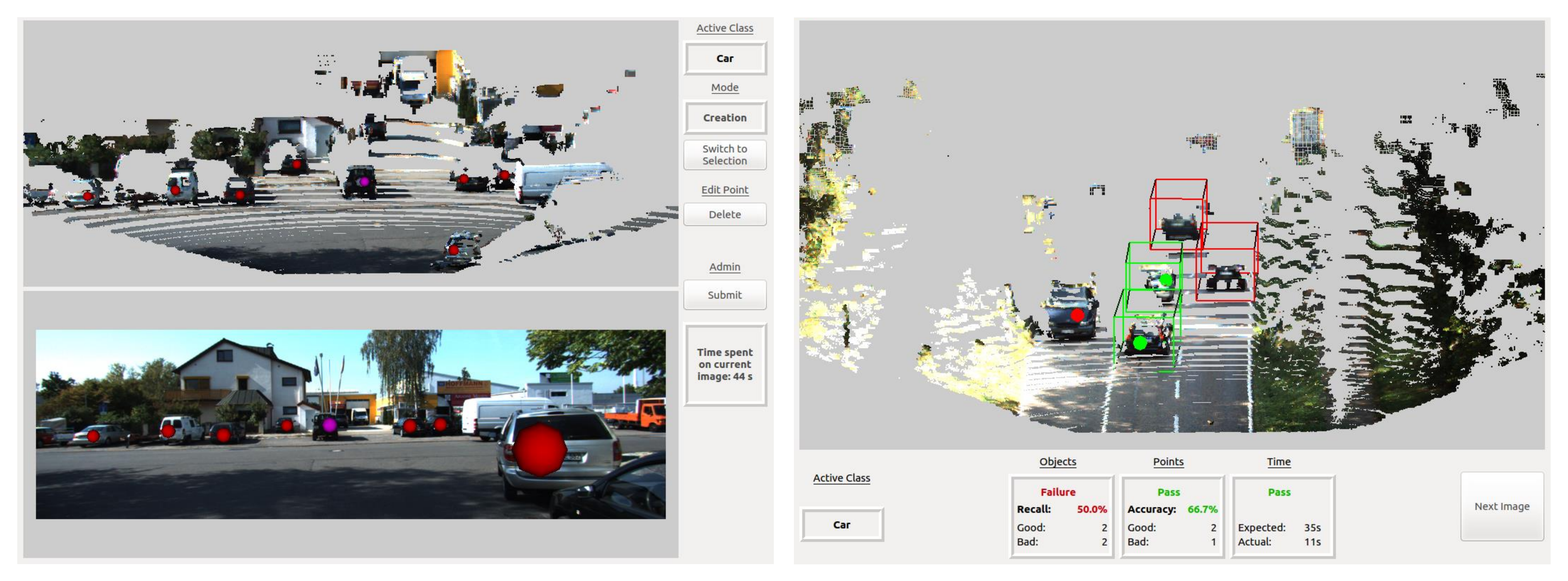}
\end{center}
\caption{\textbf{Left:} The labelling GUI used by annotators to provide click labels for each object in the scene. The \textit{purple} click signifies the current point selected by the user. \textbf{Right:} The annotator assessment window shows the annotators correct annotations in green and erroneous or missed annotations in red.} 
\label{figure:gui}
\end{figure*}

Click collection is performed in 3D space, where each click is represented as an $(x,y,z)$ tuple. Users are presented with a 3D colourized point cloud, which is generated by running depth completion \cite{ku2018depth} on the LIDAR pointcloud, and are asked to click on all instances belonging to a single class. We restrict labeling to a single class to allow the users to harness their knowledge of the scene to find all object class instances, and to minimize task switching. To further ease the annotation task, scenes are presented in temporal sequence allowing the user to carry knowledge of the previous scene on to the next. As our focus is solely on 3D labels, we mandate that the clicks are collected in 3D space. This requirement forces the annotator to be fully aware of the structure of the objects and that the click has properly been placed on it. In the case of poor visibility, such as heavily shaded areas, the object's structure will still be distinguishable in the 3D space. To allow efficient annotation, we design a custom user interface based on the process described so far (Fig. \ref{figure:gui}). \\

\noindent\textbf{Annotator Training}
To maintain high-quality results, annotators are initially given a set of instructions on how to properly perform the 3D annotation task.  The annotators are then provided with a training sequence comprising of 5 scenes in which the ground labels of all objects are known. The annotators are required to label these 5 scenes and achieve a minimum level of recall and precision per scene. This is done under a time constraint relative to the number of objects in the scene. The max allowable time for a scene is computed as $T_{max} = N \times T_{object} + T_{scene}$, where $N$ is the number of objects in the scene, $T_{object}$ is the time allocated for clicking each object and $T_{scene}$ is the time allocated to initially scan and understand the whole scene. Both $T_{object}$  and $T_{scene}$ are set to $7$ seconds. When the annotators complete the annotation of these 5 scenes, they are provided with a review window as shown in Figure \ref{figure:gui}. The review window displays each of the annotated scenes, showing the position of the annotations and the ground truth bounding boxes of objects in the scene. Clicks located inside ground truth bounding boxes are displayed in green, while those that are outside are displayed in red. Recall, precision, and the time taken to annotate the scene are also provided for the annotator. The annotator must attain a recall of $0.8$, a precision of $0.6$, and complete all scenes within the allocated time for them to pass the training. Otherwise, another set of 5 scenes are provided and the annotator is required to attempt a new training sequence. Annotators can repeat training as needed until they pass, allowing for simultaneous annotator training and ability validation. This removes the need for an explicit annotator qualification test usually present in state-of-the-art methods \cite{papadopoulos2016we, papadopoulos2017training, papadopoulos2017extreme}.

\subsection{Annotating Frames}
Once the annotator passes the training stage, they are permitted to begin annotating scenes. Scenes are bundled in batches of $20$, with the annotator required to label instances of \textbf{one} object class per batch. This bundling procedure was shown to prevent task switching \cite{papadopoulos2017training}, decreasing the annotator response time and increasing their accuracy.\\  

\noindent \textbf{Quality Assurance:} In order to maintain accurate results during the annotation process, one golden question scene---for which we have the ground truth---is appended to each batch at a random location. Annotators are unaware of the existence of these scenes, and are tested for the same passing criteria as annotator training. If the annotator fails, labels in the batch are discarded and the annotator has to retake training to continue. If the annotator passes, annotation in the batch are committed to the click database and a new batch seamlessly begins. This methodology insures that if the quality of annotations deteriorates, then the labels are not saved.

\section{3D Instance Segmentation and Bounding Box Estimation}
After click annotations have been collected and stored in a click database, we design a neural network architecture to perform instance-level object segmentation followed by 3D bounding box estimation. We base our model on the state-of-the-art F-PointNet architecture \cite{qi2017frustum} for 3D object detection. Our proposed model contains three submodules, an instance segmentation network, a centroid regression network, and a bounding box estimation network that are run in succession. The instance segmentation module takes click annotations and outputs points in the 3D scene belonging to the object associated with each click. The centroid regression network takes these points as input to estimate the object's bounding box centroid. The final subnetwork, the bounding box estimation network, takes the 3D points and the estimated box centroid, and outputs the final object bounding box parameterized by its centroid position, height, width, length and orientation. The following section describes each subnetwork in more detail.

\subsection{Instance Segmentation Network:} To perform instance level segmentation on point clouds, we modify the point cloud semantic segmentation network of PointNet\cite{qi2016pointnet}  to output binary classes. The network takes as an input 3D points contained in a $K \times K \times K$ 3D volume centered at a click provided by the annotator, where $K$ is a class specific scale parameter. The points are centered around the click by subtracting the click coordinate from every point coordinate. The network is required to assign each point in the volume a binary label of ${1, 0}$ indicating if the belongs to the object or to the background. Since we only care about human annotator time and not compute time, we use a separate model for each class in our data. This was empirically seen to provide better segmentation results when compared to a single model for all required classes.

Since we do not have ground truth for the 3D instance segmentation task in the KITTI dataset, we consider all points inside the 3D object bounding boxes ground truth as our instance segmentation ground truth. Furthermore, we do not have click annotation for the training set on KITTI. Instead, random points inside the object instances are picked to simulate human clicks on the training set. This random selection also serves as a data augmentation mechanism to provide more robust instance segmentation results.

\subsection{Centroid Regression and Bounding Box Estimation Networks}
The output of the instance segmentation network is then provided as input to a 3D spatial transformer network (T-Net) \cite{qi2016pointnet} that is used to regress an approximate bounding box centroid. Similar to the residual centroid estimation of F-PointNet \cite{qi2017frustum}, the box centroid is regressed from the centroid of the instance point cloud. The points belonging to the instance are then normalized by subtracting the coordinates of the estimated centroid from their coordinates. These points are then given as input to the third subnetwork, the bounding box estimation network. For bounding box estimation, we use the classification variant of PointNet, but replace the classification output with bounding box regression. Similar to \cite{qi2017frustum}, the bounding box centroid is regressed again in a residual fashion from the previous centroid regression results. For bounding box extent regression, we follow the discrete-continuous loss definition described in F-PointNet. The model employs 4 templates (2 car, 1 cyclist and 1 pedestrian) based on the mean size of each object class from the training data split. Unlike traditional 3D object detectors \cite{ku2017joint}, where every template is regressed and pruned via non-maximum suppression, the following model generates single best solution for each instance. This is done by classifying which template fits the data best, and then regressing the offset in height, width, length and orientation from that template. Ground truth templates are chosen as ones that have the greatest 3D Intersection-Over-Union with the ground truth 3D bounding box. For more information on this hybrid regression approach, we refer the reader to \cite{MousavianCVPR17,qi2017frustum}.

\subsection{Training The Subnetworks}

For training the segmentation model, we used cross-entropy loss on the output of each point. For centroid and bounding box regression, we use the smooth L1 (Huber) loss.  

For all the networks, we use the ADAM optimizer with an initial learning rate of $0.01$ and exponential learning schedule that decays the initial learning rate by $0.7$ every $12,500$ iterations. For regularization, we use dropout at $0.7$ keep probability, and early stopping to choose the best model weights. All other parameters are kept as in the default implementation provided in \cite{qi2016pointnet}.
\section{Experiments and Results}
\label{exp}
We evaluate the performance of our annotation scheme using two criteria, the time taken for generating annotations and the quality of the generated annotations. We use the KITTI object detection dataset as example data to be labelled. We use the training-validation split of \cite{cvpr17chen}, resulting in an 1:1 ratio for training and validation set. All of our neural network training is performed on the $3,712$ frames from the training set, while our annotation procedure is tested on the $3,769$ scenes of the validation set. To prove generalization capabilities, we also test our annotation method on $300$ frames collected by our autonomous vehicle, the Autonomoose. We refer to this data as the "Autonomoose" data.
\begin{figure}[t] 
\begin{center}
\includegraphics[width=\columnwidth]{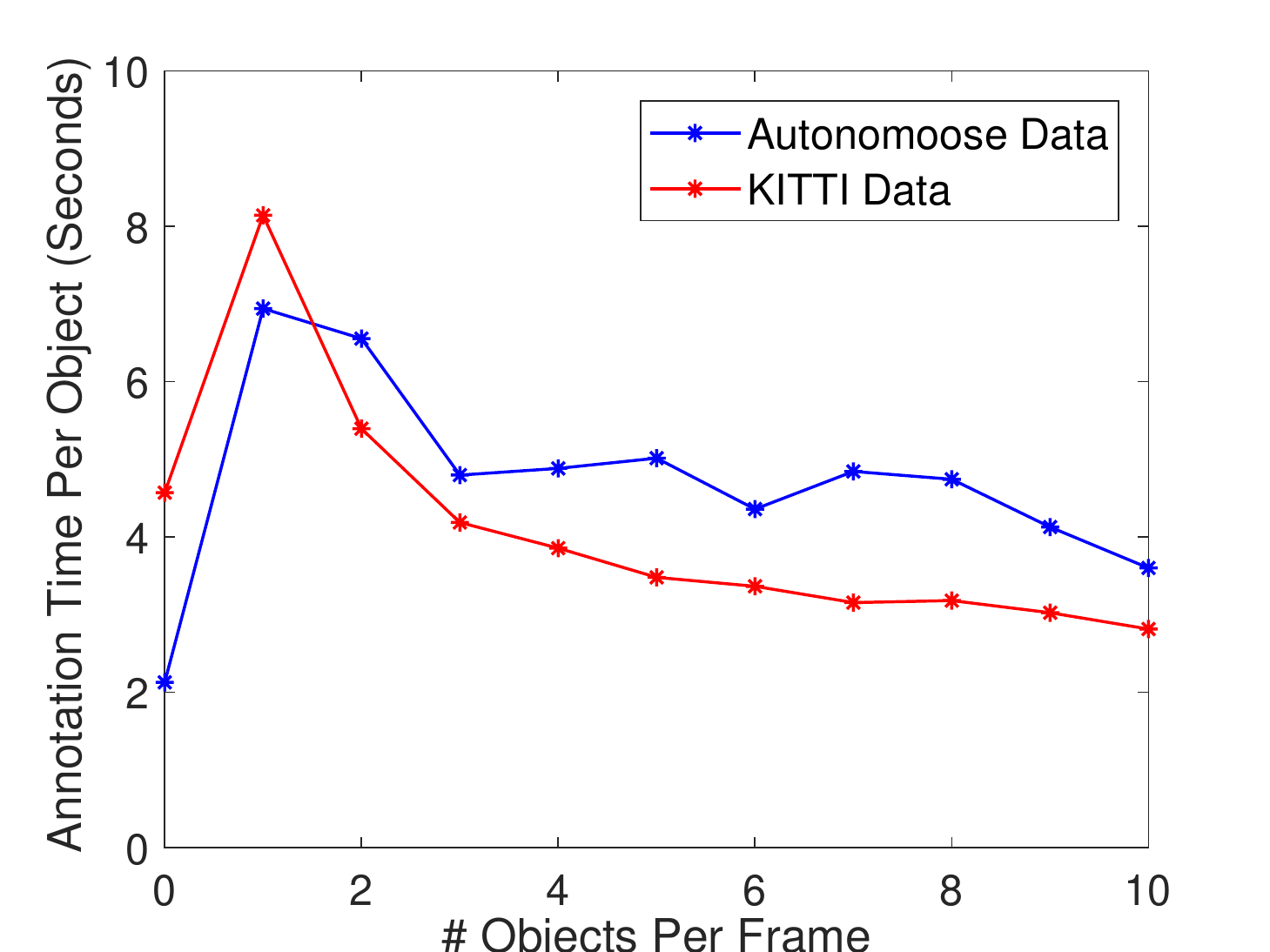}
\caption{Annotation time per object (in seconds) vs the number of objects in the scene plots for annotating the KITTI validation set and the Autonomoose data using the proposed annotation scheme.} 
\label{figure:ano_time}
\end{center}
\end{figure}

\begin{figure*}[t] 
\begin{center}
\includegraphics[width=0.9\textwidth]{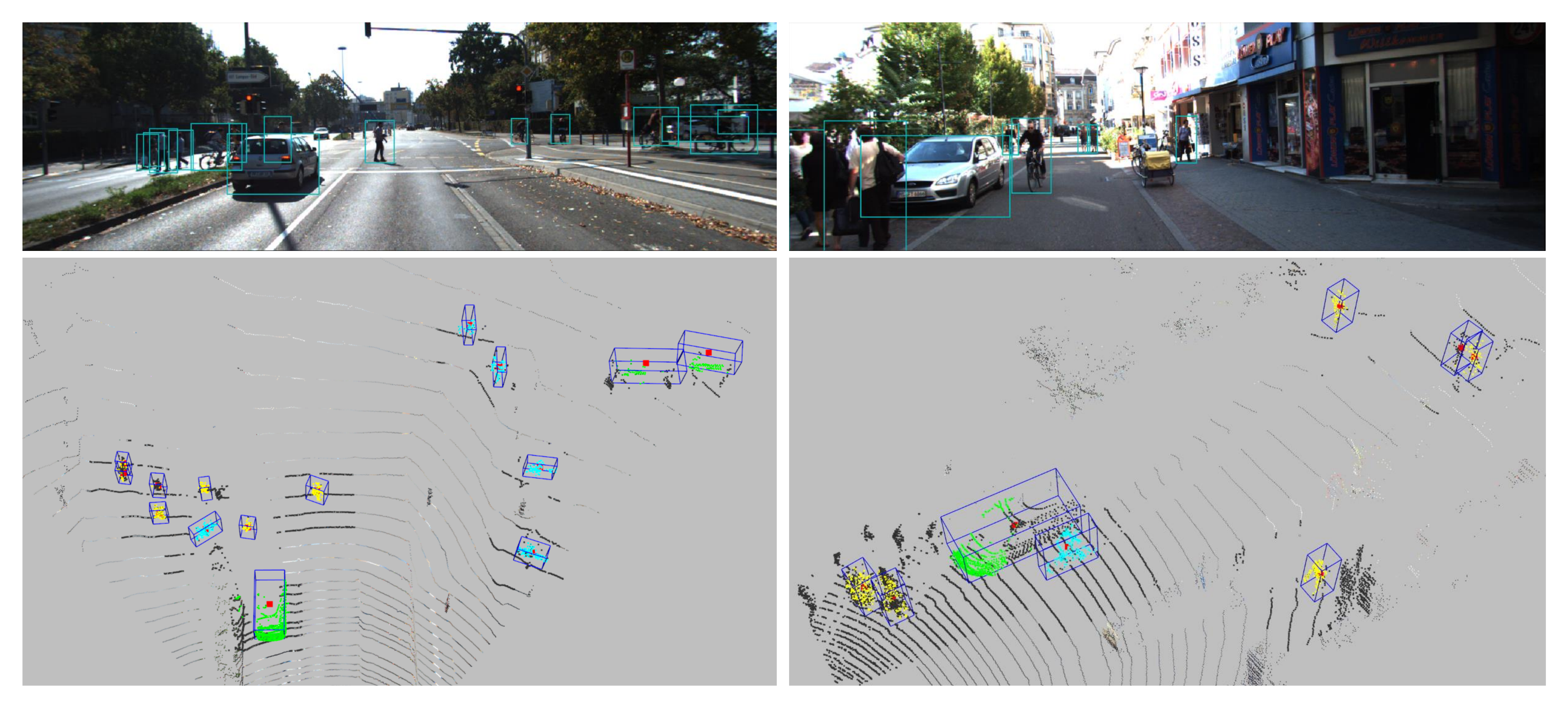}
\caption{Qualitative results of the three subnetworks of the proposed system. Instance level segmentation results for the three classes are shown in \textit{green} for cars, \textit{yellow} for pedestrians and \textit{cyan} for cyclists. Centroid estimation is shown in \textit{red} and the final 3D bounding box in \textit{blue}.  } 
\label{figure:res_kitti}
\end{center}
\end{figure*}

\subsection{Efficiency and Quality of Our Proposed Annotation Scheme}
In total, $58,832$ seconds were required to annotate all $15,996$ car instances in the $3,769$ frames from the KITTI validation set. The car class is used exclusively, as other classes comprise a very low percentage of the validation data. For labeling the $300$ frames collected with the Autonomoose, annotation required $6,954$ seconds, just under two hours, for $1,152$ car and $138$ pedestrian labels. The proposed annotation scheme requires around $3.7$ seconds to annotate a single bounding box on the KITTI dataset, and  around $6$ seconds to annotate a single bounding box on our data. At roughly $114$ seconds, the only published 3D object annotation scheme \cite{song2015sun}, takes at least $19\times$ longer per bounding box. We have noticed that generating annotations on our data takes on average around $1.3$ additional seconds per label. This can be attributed to the lower quality LIDAR and camera images provided by our data when compared to the KITTI dataset, which requires annotators to spend more time searching for all object instances in the scene. 

However, studying the average time is not enough to fully describe the efficiency of our annotation scheme. Figure \ref{figure:ano_time} shows a plot of the annotation time per object vs the number of instances in the scene. It can be noted that for both KITTI data and the Autonomoose data, annotators tend to spend less time per object as the number of objects in the scene increase. This implies that annotation speed increases as the time spent observing the scene increases, making our approach efficient for multi-object label generation.

Finally, we use the validation set ground truth bounding boxes to test precision and recall of the click collection scheme. We achieve a $96.5\%$ recall on the KITTI validation set, with a precision of $77.06\%$. The main reason for missed instances was the lack of sufficient points belonging to these instances in the point cloud. This can be remedied by using higher resolution point clouds, attained either through the use of more advanced hardware or temporal concatenation. Additionally, we believe higher resolution point clouds will help boost the precision of annotations, as more details can be seen in the point cloud to identify classes. 

\subsection{Instance Level Segmentation Performance}
To measure how well we are able to generate instance segmentation labels, we use the average instance-level IOU in 3D, defined as:
\begin{align} 
 iIOU =  \frac{\mathbb{P} \cap \mathbb{P}^*}{\mathbb{P}  \cup \mathbb{P}^*},
\end{align}
where $\mathbb{P}$ is the set of predicted points and $\mathbb{P}^*$ is the set of ground truth points of each instance.
Table \ref{table:results} provides the results of instance segmentation on the three classes of the KITTI dataset. It can be seen that we have an IOU larger than $80\%$ for the three largest classes of the KITTI dataset. To showcase the quality of the instance masks generated, we present qualitative results in Figure \ref{figure:res_kitti}. It can be seen that points on most object instances are differentiated well from the background. 

Note that for pedestrians and cyclists, we use random points inside the ground truth 3D bounding box to train and evaluate as we did not collect point clicks for these classes. For the car class, annotator clicks are used.

\begin{table*}[t]
    \begin{center}
        \resizebox{\textwidth}{!}{
            \begin{tabular}{l || c  c  c c c } \toprule
                 Class & Number of Instances&  i-IoU &  Centroid Distance Error [m] &  3D Box IoU &   Average Precision (3D) \\
                \midrule
                \textbf{Car}     & 14,318  & 0.84   &  $\pm0.23$ &  0.70 & 88.33 \\\midrule
                \textbf{Pedestrian} & 2,280  &  0.88  & $\pm0.13$ &  0.47 & 88.73    \\ \midrule
                \textbf{Cyclist}   & 893 &  0.82   &  $\pm0.22$  &  0.56 & 87.31\\
                \bottomrule
            \end{tabular}
        }
    \end{center}
    \caption{Qualitative performance evaluation of the subnetworks in our annotation scheme. All metrics are averages across all instances belonging to a single class.}
    \label{table:results}
\end{table*}

\subsection{Bounding Box Estimation}
To test the quality of our bounding box estimation scheme, we use the average box 3D Intersection-Over-Union(IOU) metric. IOU decreases as the boxes are further away from the ground truth boxes in size, position, and orientation. Furthermore, we treat our annotation scheme as a standard 3D detector and evaluate its performance with the Average Precision metric (AP) in 3D \cite{geiger2012we} at $0.5$ IOU threshold for the \textit{car} class and $0.25$ IOU threshold for the \textit{pedestrian} and \textit{cyclist} classes in 2D and 3D, respectively. Table \ref{table:results} presents the results of bounding box estimation on the three classes of the KITTI dataset. Our proposed annotation scheme achieves an $88.33\%$, $88.73\%$ and $87.31\%$ 3D AP for the \textit{car}, \textit{pedestrian}, and \textit{cyclist} classes, respectively. The average distance error in centroid estimation however increases from $0.12$ meters for pedestrians to $0.23$ and $0.22$ meters for the car and cyclist classes. We noticed that as objects increase in size, it becomes more challenging for our deep model to perform centroid estimation from incomplete point clouds. When we analyze the average 3D Box IOU, we notice that our system performs the best on the larger sized classes (\textit{car} and \textit{cyclist}). This phenomenon can be attributed to the higher sensitivity of IOU to errors as the ground truth bounding boxes become smaller.

\subsection{Generalization To Unseen Data}
\begin{figure*}[t] 
\begin{center}
\includegraphics[width=\textwidth]{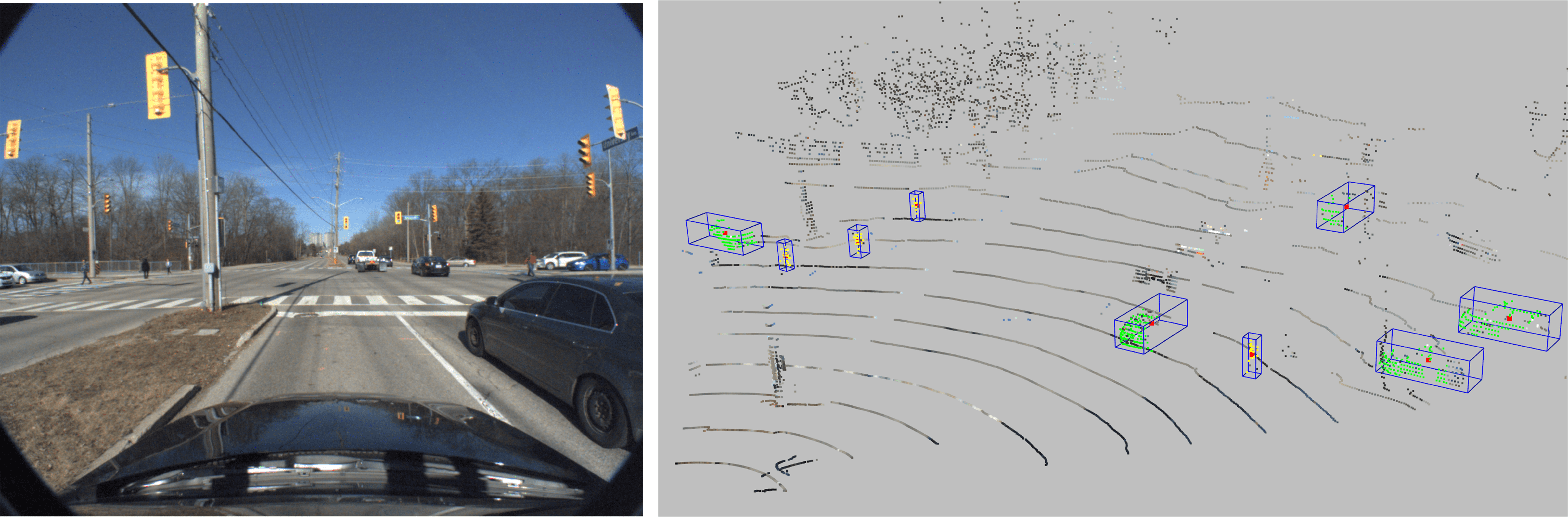}
\caption{Qualitative results of the annotation scheme on data collected by our autonomous vehicle. Instance level segmentation results for the three classes are shown in \textit{green} for cars, \textit{yellow} for pedestrians and \textit{cyan} for cyclists. Centroid estimation is shown in \textit{red} and the final 3D bounding box in \textit{blue}.} 
\label{figure:res}
\end{center}
\end{figure*}

We now test the ability of our annotation scheme to facilitate the labelling of otherwise previously unseen unlabelled data. The system expects the  unseen data to have similar point clouds to the dataset on which the system was trained. We collect the Autonomoose dataset using a VLP-32 Velodyne LIDAR and attempt to generate annotations using the proposed scheme. Using our three subnetworks, trained on the 3,712 training scenes from the Kitti dataset, we generate annotations for $300$ collected frames. Figure \ref{figure:res} shows qualitative results containing multiple cars and pedestrians from the annotated frames. It can be seen that the three subnetworks coupled with the click collection scheme generalize well to new unseen data.
\subsection{Limitations And Future Work}
Our proposed scheme is not free from limitations. First, the error in centroid estimation is still too large, resulting in a reduction in 3D IOU when compared to the KITTI dataset. Second, our bounding box regression networks require a minimum number of points on the object to be able accurately generate 3D bounding boxes. However, we argue that these two limitations are to a lesser extent generally shared with human beings when trying to estimate centroid location and bounding box extents from $2.5D$ information. Third, our annotation scheme lacks a validation procedure to filter out erroneous results from each subnetwork, which leads to cascaded errors from each subnetwork affecting the overall performance in a compound manner. Finally, we expect that a deep model specifically tailored for click annotations would provide better results than the current architecture inspired by F-PointNet. Future work will include directly incorporating clicks within the model, and incorporating a validation/retraining process to fix errors in the output of the subnetworks similar to \cite{papadopoulos2016we}.

\section{Conclusion}
In this work, we proposed a hybrid annotation scheme to create high quality 3D ground truth labels with minimal annotator effort. Our annotation scheme requires annotators to simply click on objects within a 3D LIDAR point cloud. The proposed network employs a first stage segmentation structure, using the provided annotation labels as seeds to generate accurate instance level segmentation results for each object. Through the use of a center-regression T-Net, the centroid of each object is estimated and finally, a bounding box is fit to the object in a third stage. Since the only interaction required by annotators is to provide the initial object instance clicks, the time taken to generate a bounding box for each object can be reduced to the annotation time, which is up to 30x faster than existing known methods of producing ground truth for 3D object detection. 

{\small
\bibliographystyle{unsrt}
\bibliography{label3d_bib}
}
\end{document}